\documentclass[runningheads]{llncs}

\usepackage[T1]{fontenc}
\usepackage{graphicx}
\usepackage{amsmath}
\usepackage{array}
\usepackage{multirow}
\usepackage{arydshln} % 用于绘制 \hdashline 虚线
\usepackage{makecell} % 用于表头内的优雅换行
\usepackage{placeins}
\usepackage{subcaption}
\usepackage{booktabs}

\usepackage{algorithm}
\usepackage{algpseudocode}
\usepackage{arydshln}
\usepackage{xcolor}
\usepackage{amssymb}
\usepackage{amsfonts}
\usepackage{caption}
\usepackage{cite}

\usepackage{bbding}
\usepackage{hyperref}
\hypersetup{
colorlinks=true,  % 使用文字颜色而非边框表示超链接
urlcolor=black,  % URL颜色
citecolor=blue,  % 文献引用颜色
linkcolor=blue  % 图表、公式引用颜色
}

\newcommand{\OurMethod}{Adaptive Depth Sparse Framework: Similarity-Driven Resource Allocation for Pre-Trained LLMs}
\newcommand{\suoxie}{AdaDSF}

\begin{document}

\title{\OurMethod}
\author{Yidu Wu\inst{1} \and Xiang Wang\inst{2} \and Kejie Zhao\inst{3} \and Zhangchi Wang\inst{1} \and Qinghai Guo\inst{2} \and Xiaoying Tang\inst{1}\textsuperscript{\Envelope} }
\institute{Department of Electronic and Electrical Engineering, Southern University of Science and
Technology, Shenzhen, China \\
\email{tangxy@sustech.edu.cn}
\and ACS Lab, Huawei Technologies Co., Ltd., Shenzhen, China
\and Department of Computer Science and Engineering, Southern University of Science and
Technology, Shenzhen, China
}

\authorrunning{Y. Wu, X. Wang, K. Zhao, Z. Wang, Q. Guo, and X. Tang}

\titlerunning{AdaDSF: Similarity-Driven Resource Allocation for Pre-Trained LLMs}

\maketitle

\begin{abstract}
Large language models (LLMs) achieve strong generation and reasoning performance, but the Transformer architecture incurs high inference cost. Existing acceleration methods often rely on task-specific fine-tuning or training from scratch, increasing adaptation cost and limiting cross-task usability. We present an \emph{Adaptive Depth Sparse Framework} (\suoxie) that converts off-the-shelf pre-trained LLMs into depth-sparse models without full retraining. Our key insight is that layers contribute unequally to representation transformation, characterized by the cosine similarity between layer input and output hidden states. Based on this, \suoxie\ assigns layer-wise token retention ratios from similarity statistics, uses a lightweight router to select informative tokens at each layer, and introduces a feature-preserving alignment objective to match intermediate and final representations between sparse and dense models. On GPT-NeoX and Qwen2.5 over language modeling and commonsense reasoning, \suoxie\ substantially reduces inference FLOPs while preserving performance close to dense counterparts. Under comparable sparsity, \suoxie\ consistently yields smaller accuracy degradation than strong baselines including MoD, D-LLM, and DLO.
\end{abstract}

\section{Introduction}

Large language models (LLMs)\cite{touvron2023llamaopenefficientfoundation,team2024qwen2,black2022gpt} have become the foundation of modern language understanding and generation, with applications now extending to autonomous structured-action settings such as agentic Text-to-SQL pipelines\cite{su2026agentic}. Yet deploying LLMs in real-world settings remains challenging because Transformer inference scales linearly with depth and quadratically with sequence length, leading to high computational cost.

Existing efficient-inference work spans quantization\cite{cai2020zeroq,nagel2020up,frantar2022gptq}, knowledge distillation\cite{jiao-etal-2020-tinybert,xu2024survey}, and lightweight attention or architectural redesign\cite{katharopoulos2020transformers}. More recently, depth-sparse methods such as Mixture-of-Depths (MoD)\cite{raposo2024mixture}, D-LLM\cite{jiang2024d}, and DLO\cite{tan2025dlo} reduce computation by executing full blocks only for selected tokens or layers, demonstrating that token contributions are highly non-uniform and that conditional depth execution can significantly reduce FLOPs.

Despite these advances, current depth-sparse approaches still face practical limitations: many depend on task-specific tuning or specialized training pipelines, weakening the portability of pre-trained checkpoints; fixed or heuristic token-retention schedules may misallocate computation across layers with different transformation roles; and methods requiring substantial architectural intervention increase implementation complexity and reduce reproducibility in off-the-shelf deployment.

To address these issues, we propose \OurMethod\ (\suoxie), an adaptive depth-sparse framework that converts pre-trained LLMs into sparse variants with minimal architectural change. Our key observation is that layers contribute unevenly to representation transformation, measured via cosine similarity between layer input and output hidden states. \suoxie\ assigns layer-wise token retention ratios from this signal, applies a lightweight router to select informative tokens, and introduces a feature-preserving alignment objective to match sparse and dense representations.

Our contributions: (i) \textbf{similarity-driven depth allocation}, deriving token retention ratios from hidden-state similarity statistics so that more compute goes to layers with stronger representation transformation; (ii) \textbf{lightweight routing} via an MLP-based token router for dynamic per-layer selection without redesigning the Transformer backbone; (iii) a \textbf{feature-preserving alignment objective} that preserves intermediate and final dense-teacher representations under sparse inference; and (iv) \textbf{empirical validation} on GPT-NeoX and Qwen2.5, achieving better accuracy--efficiency trade-offs than strong depth-sparse baselines under comparable sparsity.

\section{Related Work}

\subsection{Sparse Conditional Computation in LLMs}

Sparse conditional computation has become a major direction for efficiency. Mixture-of-Experts (MoE) methods activate only a subset of experts per token, increasing capacity without proportional compute growth. Switch Transformer\cite{fedus2022switch} simplifies expert routing for stable large-scale sparse training, and GShard\cite{lepikhin2021gshard} combines conditional execution with scalable sharding. These works show that conditional activation can preserve performance with reduced effective computation, but they sparsify the expert dimension rather than depth-sparse token flow in standard dense backbones. A complementary line beyond expert sparsity has shown that representation-level supervision --- aligning intermediate features rather than only final outputs --- can substantially improve downstream performance under reduced computation~\cite{wang2026forecastingguidancerepresentationlevelsupervision}, providing a broader principle for the alignment objective we develop later.

\subsection{Depth-sparse Sparsification and Token Routing}

Orthogonal to expert sparsity, depth-sparse methods reduce redundant computation along Transformer layers. MoD\cite{raposo2024mixture} executes full blocks only on selected tokens under a fixed budget, forwarding the rest through residual paths; D-LLM\cite{jiang2024d} adds dynamic layer-level execution with reduced KV-cache usage; DLO\cite{tan2025dlo} combines skipping and expansion. Token pruning in vision transformers\cite{tang2023dynamic,si2023token,dosovitskiy2021an} provides further evidence that uniform token computation is inefficient. Beyond direct depth-sparse work, analogous dynamic-gating principles appear in adjacent settings, where regime-aware gating adapts feature importance to changing contexts~\cite{202603.2262} and graph-structured state-space models exploit instance-specific dynamics for structured prediction~\cite{11463578} --- supporting input-dependent compute allocation over static schedules.

Different from prior approaches relying on fixed or heuristic schedules, \suoxie\ derives layer-wise retention from representation-transformation statistics --- a data-driven compute allocation principle. Unlike prior adaptive depth approaches, our method requires no modifications to internal attention or layer expansion; the similarity-based retention strategy applies to existing Transformers without structural change. Feature-preserving alignment further improves robustness with only minimal modification to off-the-shelf models.

\section{Methods}

\begin{figure*}[t]
    \centering
    \includegraphics[width=0.95\textwidth]{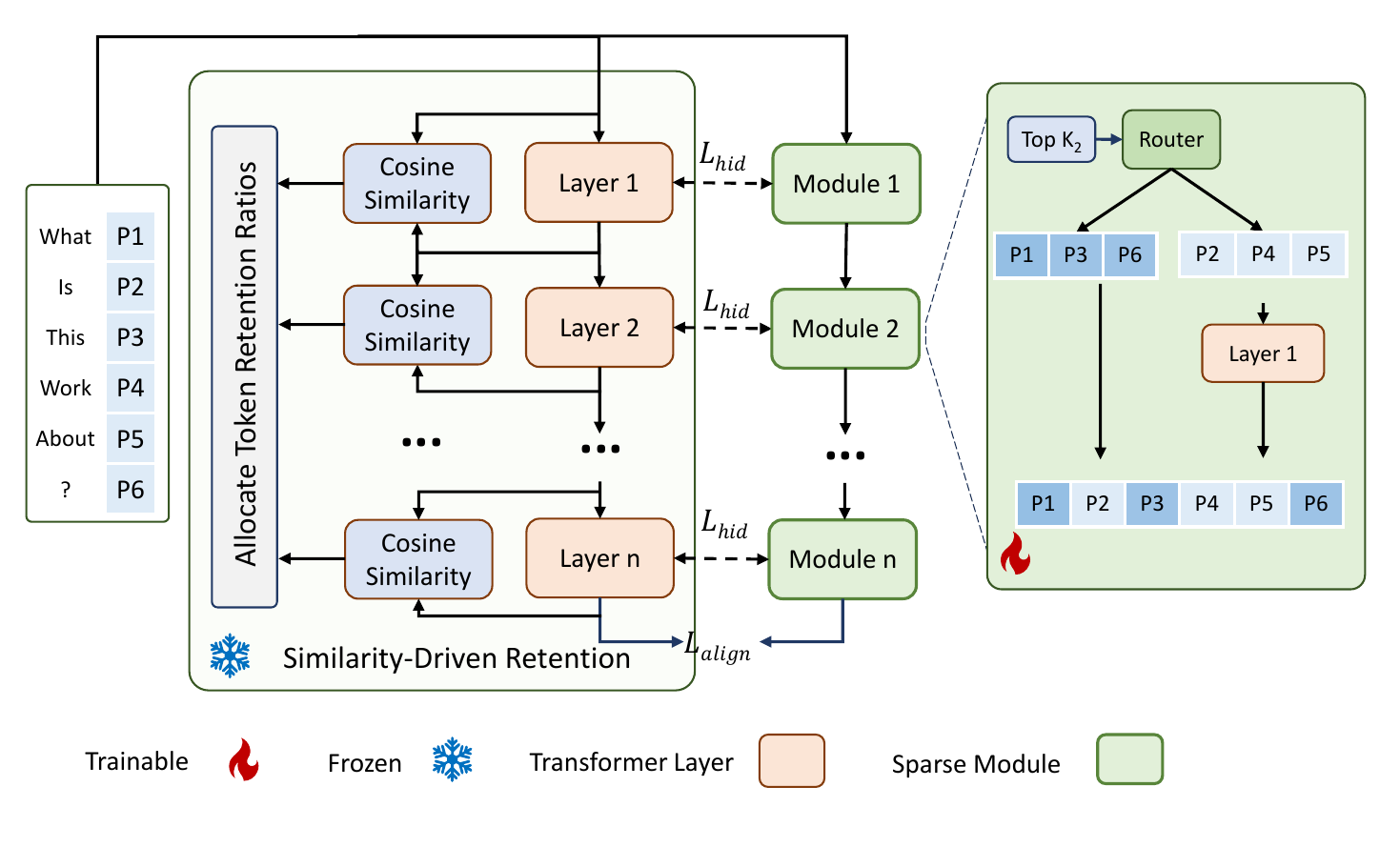}
    \caption{Overview of \OurMethod\ (\suoxie). The framework includes similarity-driven layer-wise retention allocation, lightweight token routing within sparse modules, and feature-preserving alignment training.}
    \label{fig:framwork-images}
\end{figure*}

\subsection{Preliminary: Decoder-only Transformer}

We briefly review a standard decoder-only Transformer layer. Given hidden states $H_{\text{in}}$, self-attention is
\begin{align}
\mathrm{Attention}(Q,K,V)=\mathrm{softmax}\!\left(\frac{QK^\top}{\sqrt{d_k}}\right)V,
\label{eq:attn}
\end{align}
the feed-forward network is
\begin{align}
\mathrm{FFN}(x)=W_2\!\left(\sigma(W_1x+b_1)\right)+b_2,
\label{eq:ffn}
\end{align}
and the layer output is
\begin{align}
H_{\text{out}}=\mathrm{LayerNorm}\!\left(H_{\text{in}}+\mathrm{FFN}(\mathrm{Attention}(H_{\text{in}}))\right).
\label{eq:transformer_layer}
\end{align}

\subsection{Problem Formulation and Framework Overview}

Given a pre-trained dense LLM with $L$ decoder layers, our goal is a depth-sparse variant that reduces inference FLOPs while preserving the dense teacher's behavior. Different dense layers contribute unequally to representation transformation; we quantify this via cosine similarity between layer input and output hidden states and use it to assign \emph{layer-specific token retention ratios}.

As shown in Fig.~\ref{fig:framwork-images}, \OurMethod\ has three components: (i) \textbf{similarity-driven layer-wise token retention} computes retention ratios from dense-layer similarity statistics; (ii) \textbf{a sparse module with lightweight token router} where each sparse layer contains a router and a Transformer layer, with the router selecting Top-$K$ informative tokens per the assigned ratio; and (iii) \textbf{feature-preserving alignment training} that aligns intermediate and final outputs between sparse and dense models. The design enables direct conversion of off-the-shelf checkpoints with minimal architectural changes.

\subsection{Similarity-Driven Layer-wise Token Retention}

For the $i$-th dense layer, let $x_{\text{in}}^{(i)}$ and $x_{\text{out}}^{(i)}$ be its input/output representations. We define
\begin{align}
s_i=\mathrm{CosSim}\!\left(x_{\text{in}}^{(i)},x_{\text{out}}^{(i)}\right)
=\frac{x_{\text{in}}^{(i)}\cdot x_{\text{out}}^{(i)}}{\|x_{\text{in}}^{(i)}\|\,\|x_{\text{out}}^{(i)}\|}.
\label{eq:cossim}
\end{align}
Using a calibration subset, we obtain similarity vector $s=[s_1,\dots,s_L]$.

\noindent\textbf{Step 1: Temperature-normalized weighting.}
\begin{align}
\tilde{w}_i=\exp\!\left(\frac{s_i-\max(s)}{\tau}\right),\quad
w_i=\frac{\tilde{w}_i}{\sum_{j=1}^{L}\tilde{w}_j}.
\label{eq:weight_norm}
\end{align}

\noindent\textbf{Step 2: Deviation scaling.}
\begin{align}
z_i=\beta\left(\frac{1}{L}\sum_{j=1}^{L}w_j-w_i\right),
\label{eq:deviation}
\end{align}
where $\beta>0$ controls deviation magnitude (we set $\beta=10$).

\noindent\textbf{Step 3: Sigmoid mapping to bounded ratio.}
\begin{align}
r'_i=0.05+0.9\cdot\frac{1}{1+\exp(-z_i)}.
\label{eq:sigmoid_ratio}
\end{align}

\noindent\textbf{Step 4: Global budget correction.}
Given target average retention $t$, we rescale
\begin{align}
r_i=\frac{tL}{\sum_{j=1}^{L}r'_j}\,r'_i,
\quad \text{s.t. } \sum_{i=1}^{L}r_i=tL,
\label{eq:ratio_correction}
\end{align}
so layers with larger estimated transformation receive more compute under a fixed global budget.

\subsection{Sparse Module with Lightweight Token Router}

Each sparse layer is a trainable module composed of a lightweight MLP router and one Transformer layer. For hidden states $x^{(i)}\in\mathbb{R}^{b\times s\times d}$, the router predicts token importance scores and selects Top-$K_i$ tokens, with $K_i=\lfloor r_i\cdot s\rfloor$, yielding a binary mask $m^{(i)}\in\mathbb{R}^{b\times s\times 1}$ and selected tokens $\hat{x}^{(i)}=x^{(i)}\odot m^{(i)}$. Selected tokens are processed by the Transformer layer; unselected tokens bypass computation through the residual path. The router-score pathway is kept differentiable during training to enable end-to-end optimization.

\begin{figure}[t]
    \centering
    \includegraphics[width=0.75\linewidth]{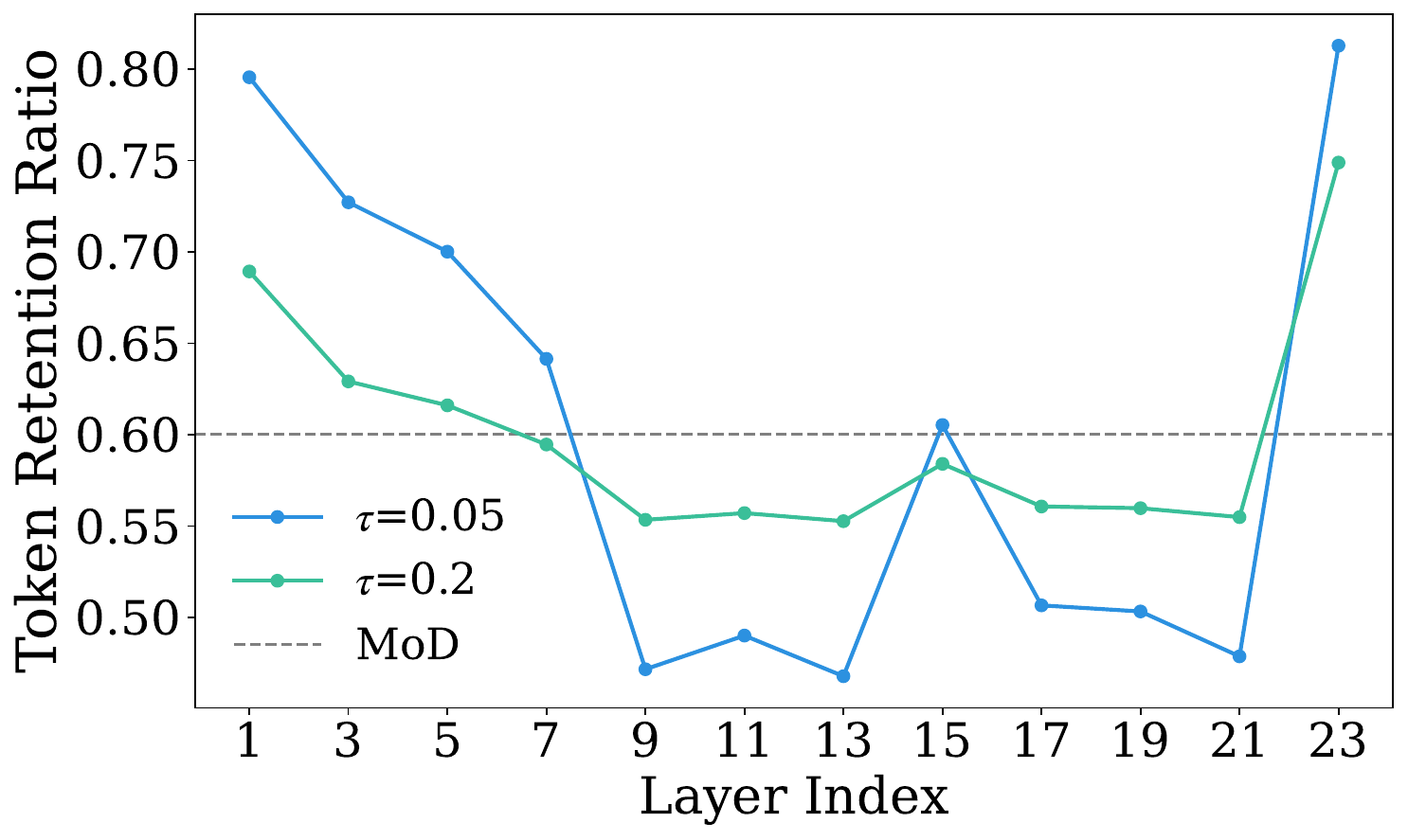}
    \caption{Layer-wise retention ratios produced by the Similarity-Driven allocation strategy.}
    \label{fig:token_keep_ratios_Qwen2.5-0.5B}
\end{figure}

\subsection{Feature-Preserving Alignment Training}

Depth sparsification introduces representation shift because some tokens skip full computation. We therefore align sparse and dense trajectories at both intermediate and output levels. This echoes recent work on instance-aware representation alignment, where per-instance correspondences capture fine-grained variation more effectively than coarse global matching~\cite{11462690}; in our setting, each token plays the role of an instance whose sparse-model representation must align to its dense-model counterpart.

\noindent\textbf{(1) Hidden-state alignment.}
For layer $l$:
\begin{align}
\mathcal{L}_{\text{hid}}^{(l)}
=
\left\|
\mathrm{Softmax}\!\left(h_{\text{sparse}}^{(l)}\right)
-
\mathrm{Softmax}\!\left(h_{\text{dense}}^{(l)}\right)
\right\|_2.
\label{eq:hid_align}
\end{align}

\noindent\textbf{(2) Output distribution alignment.}
\begin{align}
\mathcal{L}_{\text{align}}
=
\sum_{c=1}^{C}
P_{\text{dense}}(c|x)\,
\log
\frac{P_{\text{dense}}(c|x)}{P_{\text{sparse}}(c|x)}.
\label{eq:kl_align}
\end{align}

\noindent\textbf{Overall objective.}
\begin{align}
\mathcal{L}
=
\mathcal{L}_{\text{align}}
+
\frac{1}{L}\sum_{l=1}^{L}\mathcal{L}_{\text{hid}}^{(l)}.
\label{eq:total_loss}
\end{align}
This objective preserves dense behavior while enabling efficient sparse inference.

\section{Experiments}
This section provides an empirical evaluation of \suoxie\ covering settings, protocols, and analysis.

\subsection{Experimental Setup}

\paragraph{Models.}
\suoxie\ is evaluated on three representative models: GPT-NeoX-130M, Qwen2.5-0.5B, and Qwen2.5-1.5B.

\paragraph{Datasets.}
Two regimes are considered. GPT-NeoX is trained on Wikitext103, representing a pretraining scenario. Qwen2.5 models are trained on instruction-tuning data --- GenQA \cite{chen2024genqageneratingmillionsinstructions}, InfinityInstruct \cite{li2025infinityinstructscalinginstruction}, and OpenHermes2.5 \cite{OpenHermes2.5} --- to simulate settings without pretraining resources.

\paragraph{Evaluation Tasks.}
We evaluate on language modeling (Wikitext103 test set) and six commonsense reasoning benchmarks: ARC-Challenge (AC), ARC-Easy (AE) \cite{clark2018thinksolvedquestionanswering}, HellaSwag (HS) \cite{zellers2019hellaswagmachinereallyfinish}, PIQA (PI) \cite{bisk2020piqa}, WinoGrande (WG) \cite{sakaguchi2020winogrande}, and OpenBookQA (OB) \cite{mihaylov2018can}.

\paragraph{Metrics.}
Token retention ratio gives the proportion of tokens preserved at inference; computational cost is measured by normalized FLOPs (sparse over dense). For reasoning, we report accuracy with average score (Avg) and degradation relative to dense (Diff).

\subsection{Overall Performance}

\begin{table*}[t]
\centering
\subcaptionbox{Performance and FLOPs comparison of Qwen2.5-0.5B and Qwen2.5-1.5B under the MoD, DLO, and \suoxie\ methods.\label{tab:Qwen2.5-0.5Band1.5B}}{
\begin{tabular}{c|cccccccc|c}
\hline
\multirow{2}{*}{Model} & \multicolumn{8}{c|}{Accuracy (\%)} & \multirow{2}{*}{FLOPs ($\downarrow$)} \\ \cline{2-9}
 & AC & AE & HS & PI & WG & OB & Avg ($\uparrow$) & Diff ($\uparrow$) \\ \hline
Qwen2.5-0.5B & 32.3 & 64.6 & 52.1 & 70.2 & 56.3 & 35.2 & 51.7 & 0 & 1 \\
MoD & 30.5 & 52.1 & 40.5 & 63.3 & 52.5 & 27.8 & 44.4 & -7.3 & 0.784 \\
DLO & 29.8 & 62.5 & \textbf{45.3} & \textbf{68.5} & 53.7 & 30.4 & 48.3 & -3.4 & 0.973 \\
\suoxie\ (Ours) & \textbf{30.6} & \textbf{63.2} & 44.8 & 67.5 & \textbf{55.1} & \textbf{33.4} & \textbf{49.1} & \textbf{-2.6} & 0.785 \\ \hdashline
Qwen2.5-1.5B & 44.9 & 75.3 & 67.7 & 75.5 & 63.2 & 40.8 & 61.2 & 0 & 1 \\
MoD & 32.8 & 56.2 & 48.9 & 66.0 & 52.5 & 29.2 & 47.6 & -13.6 & 0.901 \\
DLO & \textbf{40.5} & 70.6 & 54.2 & \textbf{74.5} & 60.3 & \textbf{40.0} & 56.6 & -4.6 & 0.984 \\
\suoxie\ (Ours) & 39.4 & \textbf{73.5} & \textbf{58.6} & 73.1 & \textbf{61.6} & 38.2 & \textbf{57.4} & \textbf{-3.8} & 0.901 \\ \hline
\end{tabular}
}
\subcaptionbox{Performance of Qwen2.5-0.5B under different token retention ratios.\label{tab:Qwen2.5-0.5Bunderdifratio}}{
\begin{tabular}{cc|cccccccc|c}
\hline
\multirow{2}{*}{Ratio} & \multirow{2}{*}{Model} & \multicolumn{8}{c|}{Accuracy (\%)} & \multirow{2}{*}{FLOPs ($\downarrow$)} \\ \cline{3-10}
 & & AC & AE & HS & PI & WG & OB & Avg ($\uparrow$) & Diff ($\uparrow$) \\ \hline
100\% & Qwen2.5-0.5B & 32.3 & 64.6 & 52.1 & 70.2 & 56.3 & 35.2 & 51.7 & 0 & 1 \\ \hdashline
\multirow{3}{*}{90\%} & MoD & \textbf{32.5} & 53.4 & 45.1 & 65.7 & 53.8 & 31.0 & 46.9 & -4.8 & 0.889 \\
 & DLO & 31.8 & \textbf{65.7} & 44.3 & 67.5 & 54.2 & \textbf{33.4} & \textbf{49.5} & \textbf{-2.2} & 1.110 \\
 & \suoxie\ (Ours) & 31.6 & 64.2 & \textbf{45.7} & \textbf{68.1} & \textbf{54.8} & 32.0 & 49.4 & -2.3 & 0.889 \\ \hdashline
\multirow{3}{*}{80\%} & MoD & 30.5 & 52.1 & 40.5 & 63.3 & 52.5 & 27.8 & 44.4 & -7.3 & 0.784 \\
 & DLO & 29.8 & 62.5 & \textbf{45.3} & \textbf{68.5} & 53.7 & 30.4 & 48.3 & -3.4 & 0.973 \\
 & \suoxie\ (Ours) & \textbf{30.6} & \textbf{63.2} & 44.8 & 67.5 & \textbf{55.1} & \textbf{33.4} & \textbf{49.1} & \textbf{-2.6} & 0.785 \\ \hdashline
\multirow{3}{*}{70\%} & MoD & 26.6 & 47.4 & 39.9 & 63.3 & 49.2 & 27.0 & 42.2 & -9.5 & 0.684 \\
 & DLO & 28.4 & 58.3 & 40.6 & 64.4 & 53.5 & \textbf{32.7} & 46.3 & -5.4 & 0.840 \\
 & \suoxie\ (Ours) & \textbf{29.7} & \textbf{61.7} & \textbf{42.3} & \textbf{67.6} & \textbf{53.9} & 31.2 & \textbf{47.7} & \textbf{-4.0} & 0.684 \\ \hline
\end{tabular}
}
\caption{(a) Performance and FLOPs comparison for Qwen2.5 models under different sparse methods. (b) Performance under varying token retention ratios.}
\label{result:Qwen2.5}
\end{table*}

\begin{table}[htbp]
  \centering
  \renewcommand{\arraystretch}{1.15}
  \begin{tabular}{c|c|c|c}
    \hline
    Ratio & Model & \makecell{Wikitext103 PPL ($\downarrow$)} & FLOPs ($\downarrow$) \\ \hline
    100\% & GPT-Neox & 17.9 & 1.000 \\ \hdashline
    \multirow{4}{*}{90\%}
    & MoD           & 19.9          & 0.885 \\
    & D-LLM         & 1955.3          & 1.020 \\
    & DLO           & \textbf{18.3} & 1.110 \\
    & AdaDSF (Ours) & 18.5          & 0.886 \\ \hdashline
    \multirow{4}{*}{80\%}
    & MoD           & 21.6          & 0.778 \\
    & D-LLM         & 2019.0          & 0.886 \\
    & DLO           & 19.6          & 0.964 \\
    & AdaDSF (Ours) & \textbf{18.9} & 0.787 \\ \hdashline
    \multirow{4}{*}{70\%}
    & MoD           & 24.0          & 0.678 \\
    & D-LLM         & 2078.0          & 0.759 \\
    & DLO           & 21.3          & 0.827 \\
    & AdaDSF (Ours) & \textbf{19.9} & 0.680 \\ \hline
  \end{tabular}
  \caption{Performance of AdaDSF, MoD, and D-LLM Methods on the Wikitext103 dataset under Different Retention Ratio.}
  \label{tab:GPT_results}
\end{table}

\subsubsection{Experiments on Qwen2.5}
Table~\ref{result:Qwen2.5} reports performance of different depth-sparse frameworks on Qwen2.5-0.5B and Qwen2.5-1.5B. All test-task samples are excluded from training; since D-LLM is not applicable, comparisons cover MoD and DLO. The proposed method consistently outperforms both on either model, with average performance closer to the dense baseline --- indicating stronger generalization. The advantage remains stable across token retention ratios (Table~\ref{tab:Qwen2.5-0.5Bunderdifratio}).

\subsubsection{Experiments on GPT-Neox}
We evaluate \suoxie\ against MoD, D-LLM, and DLO on Wikitext103 with GPT-NeoX-130M at retention ratios of 90\%, 80\%, and 70\%. Table~\ref{tab:GPT_results} shows \suoxie\ achieves the lowest PPL among sparse methods at every compression level while maintaining the lowest or near-lowest normalized FLOPs. At 80\%, \suoxie\ reaches PPL 18.9 vs.\ 21.6 (MoD), 2019 (D-LLM), 19.6 (DLO), using only 0.787$\times$ the FLOPs of dense. The severe degradation of D-LLM indicates it is not suitable for standard autoregressive pretraining; \suoxie's PPL stays close to the dense baseline (17.9), demonstrating stable, efficient sparsification under substantial token reduction.

\subsection{Ablation Studies}
We conducted ablations on GPT-NeoX-130M and Qwen2.5 across Wikitext103 and the commonsense reasoning tasks.

\subsubsection{Similarity-Based Token Retention Ratio Allocation.}
Table~\ref{tab:ablation_t} reports the effect of similarity-driven allocation: without it, PPL is 19.69; with it, the optimal PPL drops to 18.91. Variance of the retention ratio is controlled by $\tau$ --- smaller $\tau$ yields higher variance and improved performance.

\begin{table}[htbp]
  \centering
  \begin{tabular}{c|c}
  \hline
    Model                & Metric/PPL ($\downarrow$) \\  \hline
    GPT-Neox-\suoxie\ (w/o $\tau$)    & 19.69      \\
    GPT-Neox-\suoxie\ ($\tau$ =0.25)  & 19.37      \\
    GPT-Neox-\suoxie\ ($\tau$ =0.1)   & 19.15      \\
    GPT-Neox-\suoxie\ ($\tau$ =0.05)  & \textbf{18.91}      \\  \hline
  \end{tabular}
  \caption{Ablation study on the similarity-driven adaptive token retention ratio allocation strategy}
  \label{tab:ablation_t}
\end{table}

\begin{table}[htbp]
  \centering
  \begin{tabular}{c|c}
    \hline
    Loss function     & Metric/PPL ($\downarrow$) \\ \hline
    $L_{\mathrm{causal}}$            & 20.14      \\
    Ours  & \textbf{18.91}      \\ \hline
  \end{tabular}
  \caption{Ablation study on the loss design for token-dropping layers in the mixed loss}
  \label{tab:ablation_loss}
\end{table}

\subsubsection{Intermediate Layer Output Alignment.}
The joint training strategy aligns intermediate representations between dense and sparse models, allowing the router to dynamically select informative tokens while reducing skip-induced perturbation. Table~\ref{tab:ablation_loss} shows that this loss significantly improves \suoxie.

\subsection{Discussion}
The 0.8\%--4.7\% average improvement is achieved without modifying the internal architecture of the model. Compared to approaches that require structural changes, our strategy is simple, lightweight, and easy to integrate into existing pipelines --- a favorable efficiency--complexity trade-off that highlights the contribution's practical value in deployment-oriented scenarios.

\subsection{Future Work}
Although our experiments are limited to models up to 1.5B parameters, the proposed method is inherently scalable. We plan to extend evaluation to larger LLMs (7B+) in future work to further validate generality and robustness.

\section{Conclusion}
We presented \OurMethod\ (\suoxie), an adaptive depth-sparse framework that converts off-the-shelf pre-trained LLMs into efficient sparse variants with minimal architectural change. The core idea is to allocate layer-wise compute by hidden-state similarity, route tokens dynamically with a lightweight MLP, and preserve dense behavior via feature-level alignment. On commonsense reasoning, \suoxie\ consistently yields smaller performance degradation than existing depth-sparse baselines under matched sparsity, offering a better efficiency--accuracy trade-off and a practical path for adapting pre-trained dense models to resource-constrained deployment without retraining from scratch.

\section*{Acknowledgement}
This study was supported by the National Key Research and Development Program of China (2023YFC2415400); the National Natural Science Foundation of China (T2422012); the Guangdong Basic and Applied Basic Research (2024B1515020088); the High Level of Special Funds (G030230001, G03034K003); the Guangdong Key Research and Development Program (2025B1111080001); the SUSTech Fang Keng Faculty Award.

\FloatBarrier
\bibliographystyle{splncs04}
\bibliography{refs}

@misc{touvron2023llamaopenefficientfoundation,
      title={LLaMA: Open and Efficient Foundation Language Models}, 
      author={Hugo Touvron and Thibaut Lavril and Gautier Izacard and Xavier Martinet and Marie-Anne Lachaux and Timothée Lacroix and Baptiste Rozière and Naman Goyal and Eric Hambro and Faisal Azhar and Aurelien Rodriguez and Armand Joulin and Edouard Grave and Guillaume Lample},
      year={2023},
      eprint={2302.13971},
      archivePrefix={arXiv},
      primaryClass={cs.CL},
      url={https://arxiv.org/abs/2302.13971}, 
}

@article{su2026agentic,
  title={Agentic-SQL Taxonomy: A Survey of Autonomous and Interactive Text-to-SQL with LLMs},
  author={Su, Yiyun and Zhu, Huiying and Tian, Yu and Zhao, Changruo and Peng, Zujun and Liu, Yuting and Fan, Liang and Li, Baihua and Zhang, Luyan},
  year={2026}
}

@misc{wang2026forecastingguidancerepresentationlevelsupervision,
  title={Forecasting with Guidance: Representation-Level Supervision for Time Series Forecasting},
  author={Jiacheng Wang and Liang Fan and Baihua Li and Luyan Zhang},
  year={2026},
  eprint={2603.24262},
  archivePrefix={arXiv},
  primaryClass={cs.LG},
  url={https://arxiv.org/abs/2603.24262}
}

@article{202603.2262,
  doi={10.20944/preprints202603.2262.v1},
  url={https://doi.org/10.20944/preprints202603.2262.v1},
  year={2026},
  month={March},
  publisher={Preprints},
  author={Jiacheng Wang and Liang Fan and Baihua Li and Luyan Zhang},
  title={A Dynamic Factor Gating Architecture with Market Regime Awareness for Stock Return Forecasting},
  journal={Preprints}
}

@INPROCEEDINGS{11463578,
  author={Lu, Yao and Hu, Kaiyi and Zhang, Luyan},
  booktitle={ICASSP 2026 - 2026 IEEE International Conference on Acoustics, Speech and Signal Processing (ICASSP)},
  title={S3G: Stock State Space Graph for Enhanced Stock Trend Prediction},
  year={2026},
  volume={},
  number={},
  pages={4081-4085},
  doi={10.1109/ICASSP55912.2026.11463578}
}

@INPROCEEDINGS{11462690,
  author={Guo, Zixin and Zhao, Kai and Zhang, Luyan},
  booktitle={ICASSP 2026 - 2026 IEEE International Conference on Acoustics, Speech and Signal Processing (ICASSP)},
  title={InstanceRSR: Real-World Super-Resolution via Instance-Aware Representation Alignment},
  year={2026},
  volume={},
  number={},
  pages={10577-10581},
  doi={10.1109/ICASSP55912.2026.11462690}
}

@article{black2022gpt,
  title={Gpt-neox-20b: An open-source autoregressive language model},
  author={Black, Sid and Biderman, Stella and Hallahan, Eric and Anthony, Quentin and Gao, Leo and Golding, Laurence and He, Horace and Leahy, Connor and McDonell, Kyle and Phang, Jason and others},
  journal={arXiv preprint arXiv:2204.06745},
  year={2022}
}

@article{team2024qwen2,
  title={Qwen2 technical report},
  author={Team, Qwen},
  journal={arXiv preprint arXiv:2407.10671},
  year={2024}
}

@inproceedings{cai2020zeroq,
  title={Zeroq: A novel zero shot quantization framework},
  author={Cai, Yaohui and Yao, Zhewei and Dong, Zhen and Gholami, Amir and Mahoney, Michael W and Keutzer, Kurt},
  booktitle={Proceedings of the IEEE/CVF conference on computer vision and pattern recognition},
  pages={13169--13178},
  year={2020}
}

@inproceedings{nagel2020up,
  title={Up or down? adaptive rounding for post-training quantization},
  author={Nagel, Markus and Amjad, Rana Ali and Van Baalen, Mart and Louizos, Christos and Blankevoort, Tijmen},
  booktitle={International conference on machine learning},
  pages={7197--7206},
  year={2020},
  organization={PMLR}
}

@article{frantar2022gptq,
  title={Gptq: Accurate post-training quantization for generative pre-trained transformers},
  author={Frantar, Elias and Ashkboos, Saleh and Hoefler, Torsten and Alistarh, Dan},
  journal={arXiv preprint arXiv:2210.17323},
  year={2022}
}

@inproceedings{jiao-etal-2020-tinybert,
    title = "{T}iny{BERT}: Distilling {BERT} for Natural Language Understanding",
    author = "Jiao, Xiaoqi  and
      Yin, Yichun  and
      Shang, Lifeng  and
      Jiang, Xin  and
      Chen, Xiao  and
      Li, Linlin  and
      Wang, Fang  and
      Liu, Qun",
    editor = "Cohn, Trevor  and
      He, Yulan  and
      Liu, Yang",
    booktitle = "Findings of the Association for Computational Linguistics: EMNLP 2020",
    month = nov,
    year = "2020",
    address = "Online",
    publisher = "Association for Computational Linguistics",
    url = "https://aclanthology.org/2020.findings-emnlp.372/",
    doi = "10.18653/v1/2020.findings-emnlp.372",
    pages = "4163--4174"
}

@article{xu2024survey,
  title={A survey on knowledge distillation of large language models},
  author={Xu, Xiaohan and Li, Ming and Tao, Chongyang and Shen, Tao and Cheng, Reynold and Li, Jinyang and Xu, Can and Tao, Dacheng and Zhou, Tianyi},
  journal={arXiv preprint arXiv:2402.13116},
  year={2024}
}

@inproceedings{katharopoulos2020transformers,
  title={Transformers are rnns: Fast autoregressive transformers with linear attention},
  author={Katharopoulos, Angelos and Vyas, Apoorv and Pappas, Nikolaos and Fleuret, Fran{\c{c}}ois},
  booktitle={International conference on machine learning},
  pages={5156--5165},
  year={2020},
  organization={PMLR}
}

@article{raposo2024mixture,
  title={Mixture-of-depths: Dynamically allocating compute in transformer-based language models},
  author={Raposo, David and Ritter, Sam and Richards, Blake and Lillicrap, Timothy and Humphreys, Peter Conway and Santoro, Adam},
  journal={arXiv preprint arXiv:2404.02258},
  year={2024}
}

@article{jiang2024d,
  title={D-llm: A token adaptive computing resource allocation strategy for large language models},
  author={Jiang, Yikun and Wang, Huanyu and Xie, Lei and Zhao, Hanbin and Qian, Hui and Lui, John and others},
  journal={Advances in Neural Information Processing Systems},
  volume={37},
  pages={1725--1749},
  year={2024}
}

@inproceedings{
tan2025dlo,
title={{DLO}: Dynamic Layer Operation for Efficient Vertical Scaling of {LLM}s},
author={Zhen Tan and Daize Dong and Xinyu Zhao and Jianing Cai and Jie Peng and Yu Cheng and Tianlong Chen},
booktitle={First Workshop on Scalable Optimization for Efficient and Adaptive Foundation Models},
year={2025},
url={https://openreview.net/forum?id=E9Jw3IHuDH}
}

@article{fedus2022switch,
  title={Switch transformers: Scaling to trillion parameter models with simple and efficient sparsity},
  author={Fedus, William and Zoph, Barret and Shazeer, Noam},
  journal={Journal of Machine Learning Research},
  volume={23},
  number={120},
  pages={1--39},
  year={2022}
}

@inproceedings{
lepikhin2021gshard,
title={{\{}GS{\}}hard: Scaling Giant Models with Conditional Computation and Automatic Sharding},
author={Dmitry Lepikhin and HyoukJoong Lee and Yuanzhong Xu and Dehao Chen and Orhan Firat and Yanping Huang and Maxim Krikun and Noam Shazeer and Zhifeng Chen},
booktitle={International Conference on Learning Representations},
year={2021},
url={https://openreview.net/forum?id=qrwe7XHTmYb}
}

@inproceedings{tang2023dynamic,
  title={Dynamic token pruning in plain vision transformers for semantic segmentation},
  author={Tang, Quan and Zhang, Bowen and Liu, Jiajun and Liu, Fagui and Liu, Yifan},
  booktitle={Proceedings of the IEEE/CVF International Conference on Computer Vision},
  pages={777--786},
  year={2023}
}

@article{si2023token,
  title={Token-Selective Vision Transformer for fine-grained image recognition of marine organisms},
  author={Si, Guangzhe and Xiao, Ying and Wei, Bin and Bullock, Leon Bevan and Wang, Yueyue and Wang, Xiaodong},
  journal={Frontiers in Marine Science},
  volume={10},
  pages={1174347},
  year={2023},
  publisher={Frontiers Media SA}
}

@inproceedings{
dosovitskiy2021an,
title={An Image is Worth 16x16 Words: Transformers for Image Recognition at Scale},
author={Alexey Dosovitskiy and Lucas Beyer and Alexander Kolesnikov and Dirk Weissenborn and Xiaohua Zhai and Thomas Unterthiner and Mostafa Dehghani and Matthias Minderer and Georg Heigold and Sylvain Gelly and Jakob Uszkoreit and Neil Houlsby},
booktitle={International Conference on Learning Representations},
year={2021},
url={https://openreview.net/forum?id=YicbFdNTTy}
}

@misc{chen2024genqageneratingmillionsinstructions,
      title={GenQA: Generating Millions of Instructions from a Handful of Prompts}, 
      author={Jiuhai Chen and Rifaa Qadri and Yuxin Wen and Neel Jain and John Kirchenbauer and Tianyi Zhou and Tom Goldstein},
      year={2024},
      eprint={2406.10323},
      archivePrefix={arXiv},
      primaryClass={cs.CL},
      url={https://arxiv.org/abs/2406.10323}, 
}

@misc{li2025infinityinstructscalinginstruction,
      title={Infinity Instruct: Scaling Instruction Selection and Synthesis to Enhance Language Models}, 
      author={Jijie Li and Li Du and Hanyu Zhao and Bo-wen Zhang and Liangdong Wang and Boyan Gao and Guang Liu and Yonghua Lin},
      year={2025},
      eprint={2506.11116},
      archivePrefix={arXiv},
      primaryClass={cs.CL},
      url={https://arxiv.org/abs/2506.11116}, 
}

@misc{OpenHermes2.5,
  title = {OpenHermes 2.5: An Open Dataset of Synthetic Data for Generalist LLM Assistants},
  author = {Teknium},
  year = {2023},
  publisher = {HuggingFace},
  url = {https://huggingface.co/datasets/teknium/OpenHermes-2.5}
}

@misc{clark2018thinksolvedquestionanswering,
      title={Think you have Solved Question Answering? Try ARC, the AI2 Reasoning Challenge}, 
      author={Peter Clark and Isaac Cowhey and Oren Etzioni and Tushar Khot and Ashish Sabharwal and Carissa Schoenick and Oyvind Tafjord},
      year={2018},
      eprint={1803.05457},
      archivePrefix={arXiv},
      primaryClass={cs.AI},
      url={https://arxiv.org/abs/1803.05457}, 
}

@misc{zellers2019hellaswagmachinereallyfinish,
      title={HellaSwag: Can a Machine Really Finish Your Sentence?}, 
      author={Rowan Zellers and Ari Holtzman and Yonatan Bisk and Ali Farhadi and Yejin Choi},
      year={2019},
      eprint={1905.07830},
      archivePrefix={arXiv},
      primaryClass={cs.CL},
      url={https://arxiv.org/abs/1905.07830}, 
}

@inproceedings{bisk2020piqa,
  title={Piqa: Reasoning about physical commonsense in natural language},
  author={Bisk, Yonatan and Zellers, Rowan and Gao, Jianfeng and Choi, Yejin and others},
  booktitle={Proceedings of the AAAI conference on artificial intelligence},
  volume={34},
  number={05},
  pages={7432--7439},
  year={2020}
}

@inproceedings{sakaguchi2020winogrande,
  title={Winogrande: An adversarial winograd schema challenge at scale},
  author={Sakaguchi, Keisuke and Le Bras, Ronan and Bhagavatula, Chandra and Choi, Yejin},
  booktitle={Proceedings of the AAAI Conference on Artificial Intelligence},
  volume={34},
  number={05},
  pages={8732--8740},
  year={2020}
}

@article{mihaylov2018can,
  title={Can a suit of armor conduct electricity? a new dataset for open book question answering},
  author={Mihaylov, Todor and Clark, Peter and Khot, Tushar and Sabharwal, Ashish},
  journal={arXiv preprint arXiv:1809.02789},
  year={2018}
}
\end{document}